\documentclass[12pt]{nature}

\usepackage{times}

\usepackage{amssymb}
\usepackage{amsmath}
\usepackage{mathrsfs}
\usepackage{tabu}
\usepackage{array,booktabs}
\usepackage{colortbl}
\usepackage{xcolor}
\usepackage[hidelinks]{hyperref}

\usepackage{graphicx}
\usepackage{capt-of}
\usepackage{caption}
\usepackage{color}
\usepackage{float}
\usepackage{tocloft}

\usepackage[textwidth=16cm,textheight=21cm]{geometry}

\newcounter{lastnote}

\author{Mikhail Genkin\textsuperscript{1}, Owen Hughes\textsuperscript{2}, and Tatiana A. Engel\textsuperscript{1}}

\date{}

\begin{document}

\title{Learning non-stationary Langevin dynamics from stochastic observations of latent trajectories}

\maketitle

\begin{affiliations}
	\item \textsuperscript{1}Cold Spring Harbor Laboratory, Cold Spring Harbor, NY 
	\vspace{-0.5cm}
	\item \textsuperscript{2}University of Michigan, Ann Arbor, MI

	Corresponding author e-mail: engel@cshl.edu
\end{affiliations}

\textbf{
Many complex systems operating far from the equilibrium exhibit stochastic dynamics that can be described by a Langevin equation. Inferring Langevin equations from data can reveal how transient dynamics of such systems give rise to their function. However, dynamics are often inaccessible directly and can be only gleaned through a stochastic observation process, which makes the inference challenging. Here we present a non-parametric framework for inferring the Langevin equation, which explicitly models the stochastic observation process and non-stationary latent dynamics. The framework accounts for the non-equilibrium initial and final states of the observed system and for the possibility that the system's dynamics define the duration of observations. Omitting any of these non-stationary components results in incorrect inference, in which erroneous features arise in the dynamics due to non-stationary data distribution. We illustrate the framework using models of neural dynamics underlying decision making in the brain.
}


Many complex systems generate coherent macroscopic behavior that can be expressed as simple laws. Such systems are commonly described by Langevin dynamics, in which deterministic forces define persistent collective trends and noise captures fast microscopic interactions\cite{Friedrich:2011gd}. Langevin equations are used to model stochastic evolution of complex systems such as neural networks\cite{Genkin:2020hz,Wimmer:2014hn,Laing:2015gp,Millman:2010jo}, motile cells\cite{Bruckner:2019dm}, swarming animals\cite{Yates:2009cp}, carbon nanotubes\cite{Sriraman:2005ez}, financial markets\cite{Friedrich:2000gd}, or climate dynamics\cite{Hasselmann:2016fg}. While such systems can be readily observed in experiments or microscopic simulations, the analytical form of the Langevin equation usually cannot be easily derived from microscopic models or physical principles. The inference of Langevin equations from data is therefore crucial to enable efficient analysis, prediction, and optimization of complex systems.

Numerous methods were proposed for inferring Langevin dynamics from stochastic trajectories, e.g., by estimating moments of local trajectory increments\cite{Friedrich:2011gd,Ragwitz:2001ct,Garcia:2018ch,Postnikov:2019em,Bruckner:2020tk,Frishman:2020eu}. However, in many complex systems, the trajectories cannot be observed directly, but are only gleaned from a stochastic observation process that depends on the latent Langevin dynamics. For example, spikes recorded from neurons in the brain form stochastic point processes with statistics controlled by the collective dynamics of the surrounding network\cite{Ladenbauer:2019kj,Churchland:2011hda,Genkin:2020hz}. Similarly, the dynamics of a protein are observed through photons emitted by fluorescent dyes tagging the protein in single-molecule microscopy experiments\cite{Chung:2012kz,Chung:2013jv,Haas:2013gc}. The Poisson noise inherent in spike or photon observations makes the inference of the underlying Langevin dynamics challenging.

This challenge can be addressed by modeling data as a doubly-stochastic processes, in which latent stochastic dynamics drive another stochastic process modeling the observations\cite{Bishop:2006ui}. The inference with latent dynamical models is data efficient as it integrates information along the entire latent trajectory, but it may be sensitive to the data distribution. Previous work only considered the inference of latent Langevin dynamics for equilibrium systems with the steady-state data distribution\cite{Genkin:2020hz,Haas:2013gc}. Whether these methods extend to non-equilibrium systems has not been tested. Yet, all living systems and physical systems that perform computations operate far from equilibrium, where transient dynamics play a key role. The inference of non-stationary Langevin dynamics from stochastic observations remains an important open problem.

Here we present an inference framework for latent Langevin dynamics which accounts for non-equilibrium statistics of latent trajectories. We show that modeling non-stationary components is critical for accurate inference, and their omission leads to biases in the estimated Langevin forces. As a working example, we model non-stationary dynamics of neural spiking activity during perceptual decision making, a process of transforming a sensory stimulus into a categorical choice\cite{Gold:2007fo}. The inference of the underlying dynamics from spikes is notoriously hard\cite{Amarasingham:2015ju}, and analyses with simple parametric models result in controversial conclusions\cite{Latimer:2015hw,Chandrasekaran:2018dv,Zylberberg:2019el,Zoltowski:2019ig}. Our framework accurately infers the Langevin dynamics from spike data generated by competing models of decision making proposed previously. Our framework can be used with different stochastic observation processes and is broadly applicable for the inference of the Langevin dynamics in non-stationary complex systems.

\subsection{\label{sec:Framework} Inference framework.} 
\begin{figure}[!h]
\begin{center}
	\includegraphics{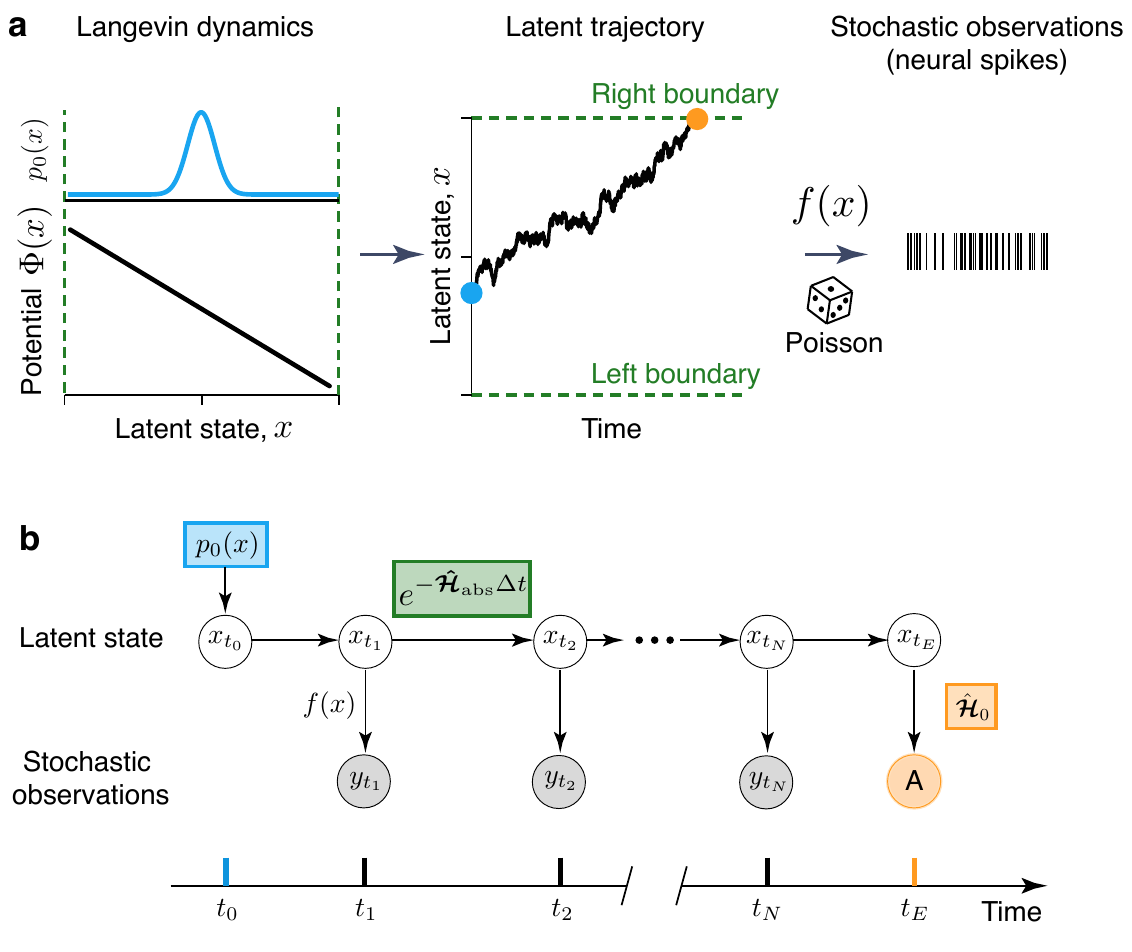}
\end{center}
\caption{\textbf{Inference framework for latent non-stationary Langevin dynamics.}
\textbf{a}, Latent dynamics are governed by the Langevin equation Eq.~(\ref{E:LatentDynModel}) with a deterministic potential $\Phi(x)$ and a Gaussian white noise with magnitude $D$. On each trial, the latent trajectory starts at the initial state $x(t_0)$ (blue dot) sampled from the probability density $p_0(x)$. When the trajectory reaches the domain boundaries (green dashed lines) for the first time, the observations can either terminate (orange dot) or continue depending on the experiment design. The latent Langevin dynamics are only accessible through stochastic observations, e.g., spikes that follow an inhomogeneous Poisson process with time-varying intensity that depends on the latent trajectory $x(t)$ via the firing-rate function $f(x)$.
\textbf{b}, Graphical diagram of the inference framework. Stochastic observations $y_{t_i}$ (grey circles) depend on the latent states $x_{t_i}$ (white circles), the arrows represent statistical dependencies. The absorption event (orange circle) indicates that observations terminate when the latent trajectory hit a boundary. The framework includes three non-stationary components: the initial state distribution $p_0(x)$ (blue box), the boundary conditions (reflecting or absorbing) for the time-propagation of latent dynamics (green box), and the absorption operator (orange box).
}\label{fig1}	
\end{figure}

We consider the inference of Langevin dynamics
\begin{equation}\label{E:LatentDynModel}
\frac{d x}{d t}=  D F(x) +\sqrt{2D}\xi(t),
\end{equation}
where $F(x)$ is the deterministic force, and $\xi(t)$ is a white Gaussian noise $\langle \xi(t) \rangle =0, \langle \xi(t)\xi(t') \rangle =\delta(t-t')$. We focus on one-dimensional (1D) Langevin dynamics representing a decision-making process on the domain $x \in [-1;1]$. In 1D, the force derives from the potential function $F(x) = - d\Phi(x)/dx$. The Langevin trajectories $x(t)$ are latent, i.e. only accessible through stochastic observations $Y(t)$. We work with observations that follow an inhomogeneous Poisson process with time-varying intensity $f(x(t))$ that depends on the latent trajectory $x(t)$ via a function $f(x)$ (Fig.~\ref{fig1}a). Poisson noise models the variability of spike generation in a neuron. 

The non-stationary data $Y(t)$ arise in non-equilibrium systems that perform computations. Such systems start their operation in a specific initial state and finish in a terminal state representing the outcome of the computation. The initial and terminal states are fundamentally different from the equilibrium state of the system. An example of such non-equilibrium computation is neural dynamics underlying perceptual decision making in the brain. Each decision process begins when a sensory stimulus is presented to a subject and terminates when the subject commits to a choice. Neural activity transiently evolves from the initial state at the stimulus onset until a choice is made, and different choices correspond to different terminal states of neural activity\cite{Gold:2007fo}. In experiment, multiple realizations of the decision process can be recorded under the same conditions, called trials. The statistics of trajectories $x(t)$ across trials differs from the steady-state distribution.

\begin{figure}[!h]
	\includegraphics{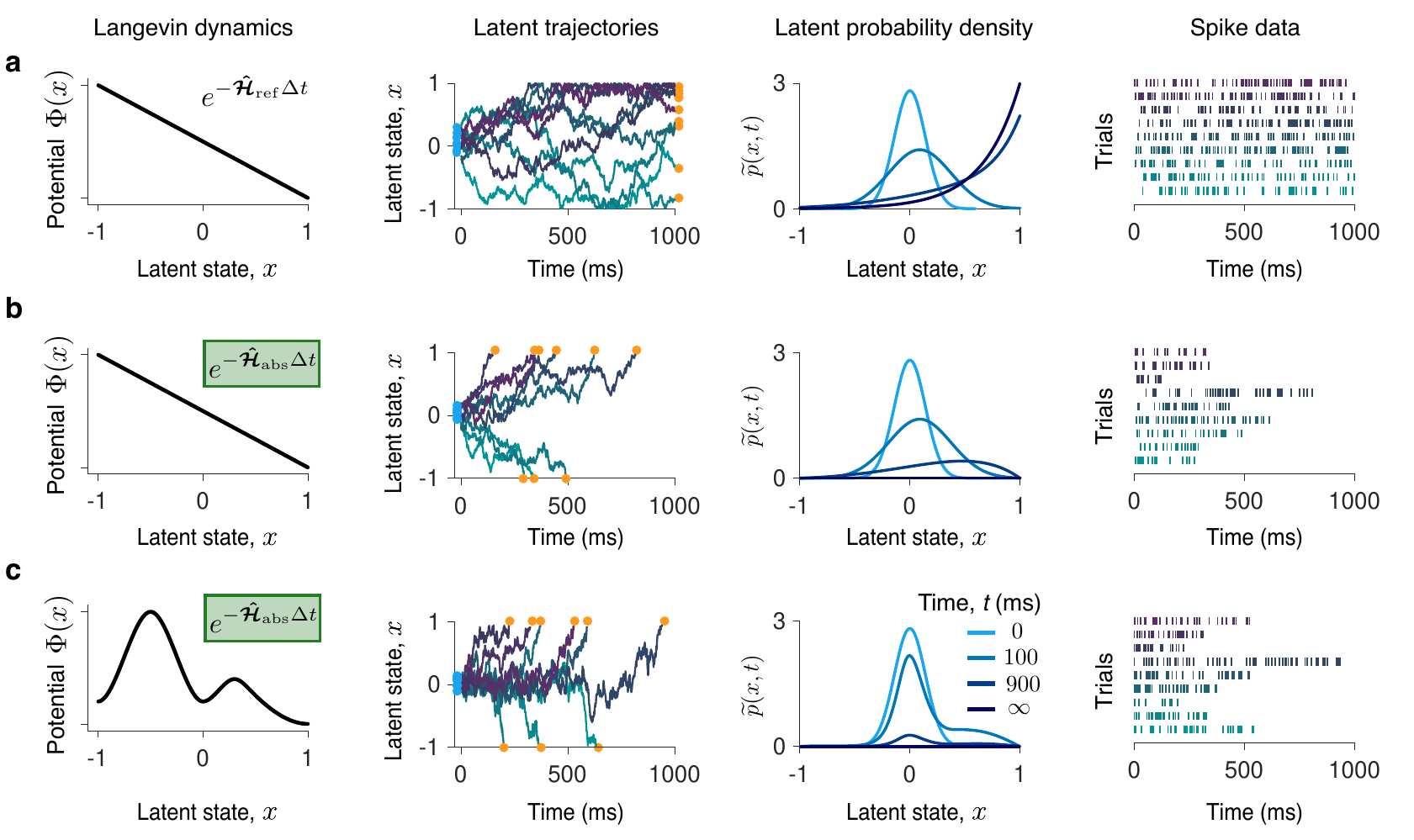}
\caption{
\textbf{Observation noise masks qualitative differences in non-stationary Langevin dynamics.}
Latent Langevin dynamics with: \textbf{a}, a linear potential and reflecting boundaries; \textbf{b}, a linear potential and absorbing boundaries; \textbf{c}, a non-linear potential and absorbing boundaries (first column). In all cases, the initial latent state $x(t_0)$ (blue dots, second column) is sampled from the same density $p_0(x)$ (third column, $t=0$). The time-propagation of the latent probability density $\widetilde{p}(x,t)$ strongly depends on the potential shape and boundary conditions (third column). With reflecting boundaries (a), the latent trajectories terminate anywhere in the latent space at the trial end, whereas with absorbing boundaries (b,c), the latent trajectories always terminate at the boundaries (orange dots, second column). These qualitative differences in the Langevin dynamics are difficult to discern from stochastic spike data (fourth column, colors correspond to the trajectories in the second column).
}\label{fig2}
\end{figure}

To model non-stationary dynamics, we introduce three components into our framework (Fig.~\ref{fig1}b). First, $p_0(x)$ models the distribution of the initial latent state at the trial start. On each trial, the latent dynamics evolve according to Eq.~(\ref{E:LatentDynModel}) from the initial condition $x(t_0)=\delta(x-x_0)$, where $x_0$ is sampled from $p_0(x)$. The distribution $p_0(x)$ is latent and needs to be inferred from data. The two other components account for the mechanism terminating the observation on each trial, which can be controlled either by the experimenter or by the system itself. In decision-making experiments, these possibilities correspond to fixed-duration or reaction-time task designs\cite{Gold:2007fo}. In a fixed-duration task, the subject reports the choice after a fixed time period set by the experimenter. Even if the neural trajectory reaches a state representing a choice (i.e. a decision boundary) at an earlier time point, the deliberation process continues. Thus, the latent trajectory can terminate at any state at the trial end (Fig.~\ref{fig2}a). In contrast, in a reaction-time task, the subject reports the choice as soon as the neural trajectory reaches a decision boundary for the first time. Thus trials have variable durations defined by the neural dynamics itself, and the latent trajectory always terminates at one of the decision boundaries at the trial end (Fig.~\ref{fig2}b,c). To model these alternative scenarios, we impose appropriate boundary conditions for the Langevin dynamics Eq.~(\ref{E:LatentDynModel}): reflecting for the fixed-duration and absorbing for the reaction-time tasks. In addition, we derive an absorption operator enforcing the trajectory termination at a decision boundary in the reaction-time task (Fig.~\ref{fig1}b).

The Poisson noise masks distinctions between different types of latent Langevin dynamics. The spike trains appear similar for dynamics with reflecting versus absorbing boundaries (Fig.~\ref{fig2}a,b), and with a linear versus non-linear potential (Fig.~\ref{fig2}b,c). A non-stationary initial state $p_0(x)$ is also not obvious in the spike trains. Distinguishing these qualitatively different dynamics based on spike data is difficult, despite the latent trajectories and the corresponding time-dependent latent probability densities $\widetilde{p}(x,t)$ (Eq.~(\ref{FP0}) in Methods) are different (Fig.~\ref{fig2}).

We infer the force potential $\Phi(x)$, the noise magnitude $D$, and the initial distribution $p_0(x)$ from stochastic spike data $Y(t)$. The data consists of multiple trials $Y(t)=\{Y_i(t)\}$ ($i=1,2,...n$), and for each trial $Y_i(t)=\{t^{i}_0,t^{i}_1,...,t^{i}_{N_i},t^{i}_E\}$, where $t_1^{i},t_2^{i},...,t_{N_i}^{i}$ are recorded spike times, and $t_0^{i}$ and $t_{\rm E}^{i}$ are the trial start and end times, respectively. We maximize the data likelihood $\mathscr{L}\left[Y(t) | \theta \right]$ with respect to $\theta=\{\Phi(x), p_0(x), D\}$. We derive analytical expressions for the variational derivatives of the negative log-likelihood, which we use to update $\theta$ using a gradient-descent (GD) algorithm\cite{Genkin:2020hz} (Methods), effectively optimizing the potential $\Phi(x)$ and $p_0(x)$ over the space of continuous functions. The likelihood calculation involves time-propagation of the latent probability density with the operator $\exp(-\boldsymbol{\hat{\mathcal{H}}}(t_i-t_{i-1})$, where $\boldsymbol{\hat{\mathcal{H}}}$ is a modified Fokker-Planck operator (Eq.~(\ref{FP}) in Methods). The operator $\boldsymbol{\hat{\mathcal{H}}}$ satisfies either reflecting ($\boldsymbol{\hat{\mathcal{H}}}_{\textrm{ref}}$, fixed-duration task), or absorbing boundary conditions ($\boldsymbol{\hat{\mathcal{H}}}_{\textrm{abs}}$, reaction-time task). The absorbing boundary conditions ensure that trajectories reaching a boundary before the trial end do not contribute to the likelihood. In addition, the absorption operator $\boldsymbol{A}$ enforces that the likelihood includes only trajectories terminating on the boundaries at the trial end $t_{\rm E}$ (Methods).

\subsection{\label{sec:NSD2} Contributions of non-stationary components to the accurate inference.}

We found that accurate inference of non-stationary Langevin dynamics requires incorporating all three non-stationary components: the initial distribution $p_0(x)$, the boundary conditions, and the absorption operator. To demonstrate how each component contributes to the accurate inference, we focus here on inferring the potential $\Phi(x)$ from synthetic data with known ground truth, assuming $p_0(x)$ and $D$ are provided (we consider simultaneous inference of $\Phi(x), p_0(x), D$ in Fig.~\ref{fig4}c). We use 200 trials of spike data generated from the model with a linear ground-truth potential and a narrow initial state distribution (full list of parameters in Supplementary Table~1). We simulated a reaction-time task, so that each trial terminates when the latent trajectory reaches one of the decision boundaries producing a non-stationary distribution of latent trajectories (Fig.~\ref{fig2}b).

\begin{figure}[!h]
	\includegraphics{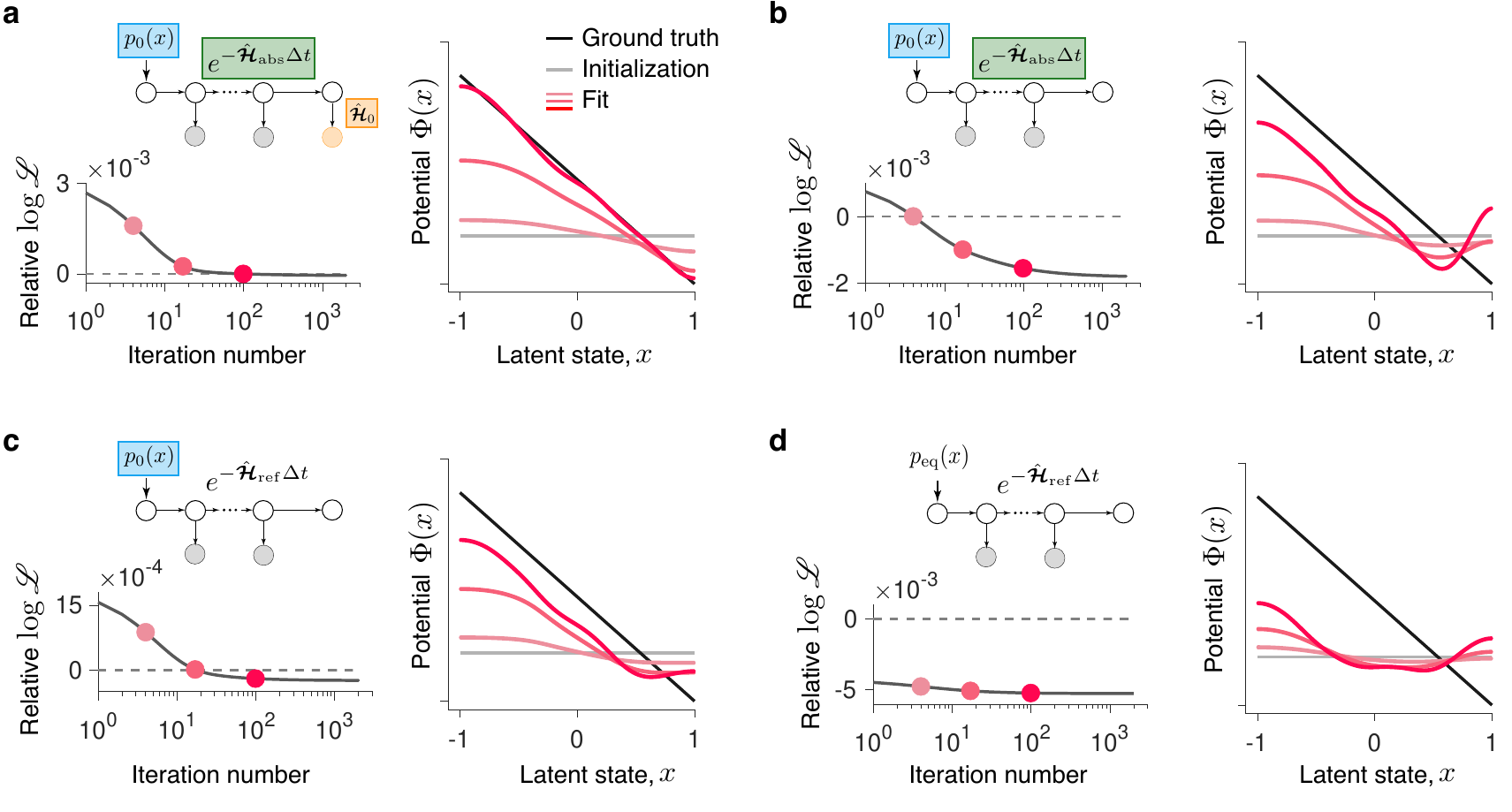}
\caption{
\textbf{Contribution of non-stationary components to the accurate inference of latent Langevin dynamics.}
The spike data are generated from the Langevin dynamics with a linear potential and absorbing boundaries (Fig.~\ref{fig2}b).
\textbf{a}, The inference incorporates the non-equilibrium initial state distribution $p_0(x)$, absorbing boundary conditions, and the absorption operator (graphical diagram, inset in the left panel). When the likelihood of the fitted model approaches the likelihood of the ground-truth model (left panel, the relative log-likelihood is $\left[ \log\mathscr{L}_{\rm gt} -  \log\mathscr{L} \right]/\log\mathscr{L}_{\rm gt}$), the inferred potential shape closely matches the ground truth (right panel, colors correspond to the iterations marked with dots on the left panel).
\textbf{b}, Same as a, but omitting the absorption operator in the inference.
\textbf{c}, Same as b, but replacing the absorbing with reflecting boundary conditions in the inference.
\textbf{d}, Same as c, but replacing $p_0(x)$ with the equilibrium density $p_{\textrm{eq}}(x)$ in the inference.
Omitting any of the non-stationary components results in artifacts in the inferred potentials.
}\label{fig3}
\end{figure}

The inference accurately recovers the Langevin dynamics from these non-stationary spike data when all non-stationary components are taken into account (Fig.~\ref{fig3}a). The GD algorithm iteratively increases the model likelihood (decreases the negative log-likelihood). Starting from an unspecific initial guess $\Phi(x)={\rm const}$, the potential shape changes gradually over the GD iterations. After some iterations, the fitted potential closely matches the ground-truth shape while the log-likelihood of the fitted model approaches the log-likelihood of the ground-truth model. The concurrent agreement of the inferred potential and its likelihood with the ground truth indicates the accurate recovery of the Langevin dynamics. At later iterations, the potential shape can deteriorate due to overfitting, and model selection is required for identifying the correct model when the ground truth is not known\cite{Genkin:2020hz}.

To reveal how each non-stationary component contributes to the inference, we replace all components one by one with their stationary counterparts and evaluate the inference quality under these modified assumptions. First, we test the importance of the absorption operator by performing the inference with the initial distribution $p_0(x)$ and absorbing boundary conditions, but omitting the absorption operator (Fig.~\ref{fig3}b). The inferred potential shows the correct linear slope, but develops a large barrier near the right boundary, where the ground-truth potential is low. This behavior arises since, without the absorption operator, the likelihood includes trajectories terminating anywhere in the latent space. The artifactual potential barrier reduces the probability flux through the absorbing boundary and therefore increases the model likelihood. Accordingly, the ground-truth potential has lower likelihood than the incorrect potential with the barrier.

Next, we test the importance of the absorbing boundary conditions using the same non-stationary data. We take into account the initial distribution $p_0(x)$, but replace the absorbing with reflecting boundary conditions in the inference (Fig.~\ref{fig3}c). The inferred potential exhibits a small barrier near the right boundary where the ground-truth potential is low. The probability density of latent trajectories in the data is vanishing at the absorbing boundaries (Fig.~\ref{fig2}b), whereas stationary dynamics with reflecting boundaries predict high probability density in the regions where the potential is low (Fig.~\ref{fig2}a). Hence, the potential barrier arises to explain the low probability density at the right boundary. Accordingly, the likelihood is higher for the model with the barrier than for the ground-truth model.

Finally, we test the importance of the initial state distribution $p_0(x)$. Using the same non-stationary data, we perform the inference with $p_0(x)$ replaced by the equilibrium distribution $p_{\rm eq}(x) \propto \exp(-\Phi(x))$ under the reflecting boundary conditions (Fig.~\ref{fig3}d). Instead of the linear slope, the inferred potential exhibits a flat shallow valley, which accounts for the high density of latent trajectories near the domain center in the data due to non-equilibrium $p_0(x)$. The equilibrium dynamics in the ground-truth potential predict lower probability density at the domain center than in the data, hence the ground truth model has lower likelihood than the inferred shallow potential. These results demonstrate that all three non-stationary components are critical for the accurate inference of non-stationary Langevin dynamics, and omitting any of them results in incorrect inference that accounts for the non-stationary data statistics by artifacts in the potential shape.

\subsection{\label{title}Discovering models of decision-making.}

To demonstrate that our framework can accurately infer qualitatively different non-stationary dynamics, we perform the inference on synthetic data generated by the alternative models of perceptual decision-making. We consider latent Langevin dynamics corresponding to the ramping and stepping models of decision making proposed previously\cite{Latimer:2015hw}. The ramping model assumes that on single trials neural activity evolves gradually towards a decision boundary as a linear drift-diffusion process, which corresponds to a linear potential with a constant slope (Fig.~\ref{fig2}b). The stepping model assumes that on single trials neural activity abruptly jumps from the initial to a final state representing a choice, which corresponds to a potential with two barriers where trajectories have to overcome one of the barriers to reach a decision boundary (Fig.~\ref{fig2}c). Distinguishing between these alternative models of decision making is difficult with the traditional approach based on parametric model comparisons\cite{Chandrasekaran:2018dv}.

\begin{figure}[!h]
	\includegraphics{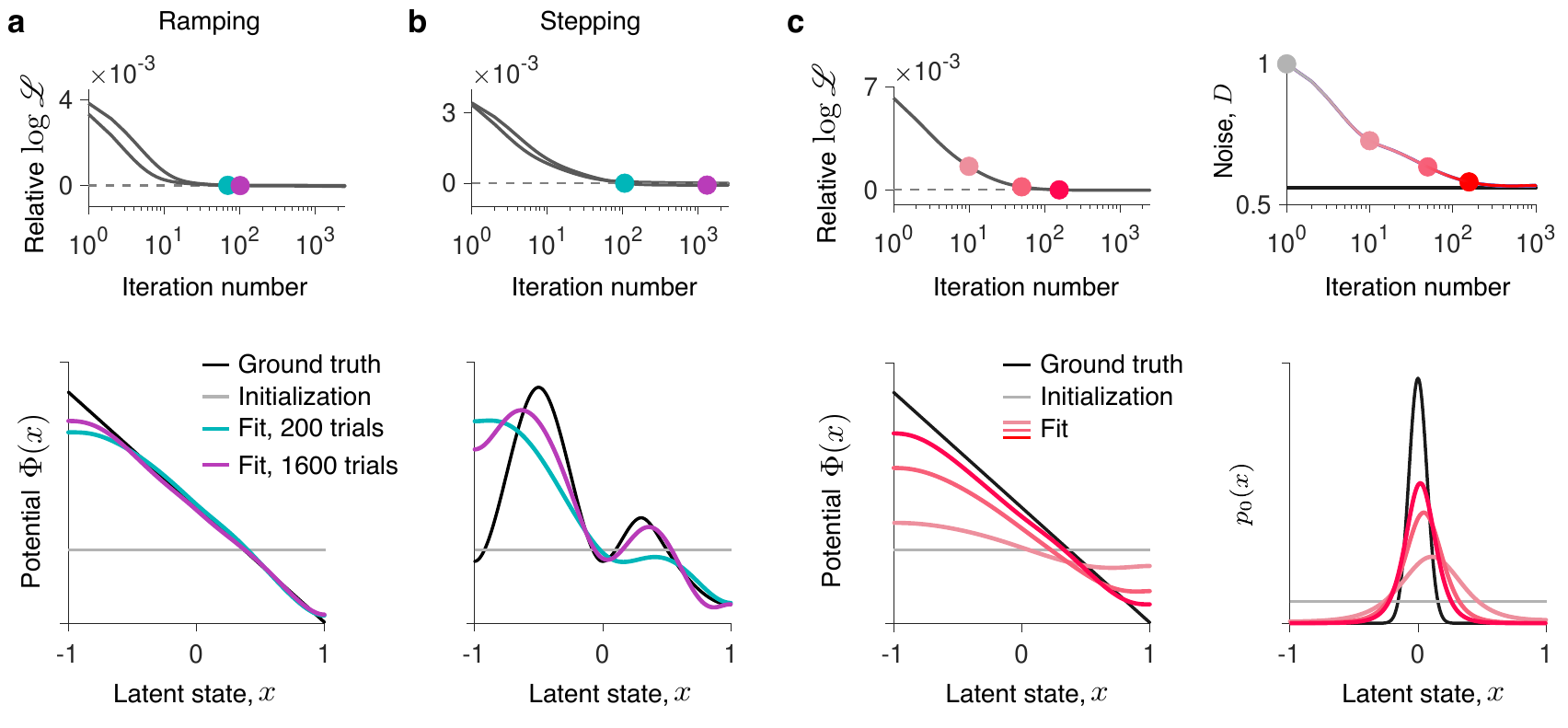}
\caption{
\textbf{Inference of decision-making dynamics and simultaneous inference of all functions governing non-stationary Langevin dynamics.}
\textbf{a}, The spike data are generated from the Langevin dynamics with a linear potential and absorbing boundaries (Fig.~\ref{fig2}b), which corresponds to the ramping model of decision-making dynamics. When the likelihoods of the fitted and ground-truth models are equal (upper panel, colored dots), the inferred potential closely matches the ground-truth potential (lower panel, colors correspond to dots in the upper panel). The inference accuracy improves with more data (teal - 200 trials, purple -  1600 trials).
\textbf{b}, Same as a, but for the spike data generated from the Langevin dynamics with a non-linear potential with two barriers and absorbing boundaries (Fig.~\ref{fig2}c), which corresponds to the stepping model of decision-making dynamics.
\textbf{c}, Simultaneous inference of the potential $\Phi(x)$, the initial state distribution $p_0(x)$, and noise magnitude $D$ from the same spike data as in a (400 trials). As the likelihood of the fitted model approaches the likelihood of the ground-truth model (upper left), all fitted components simultaneously approach the ground truth.
}\label{fig4}
\end{figure}

We generated spike data with the ramping and stepping latent dynamics in a reaction time task (Fig.~\ref{fig2}b,c). We choose the potential $\Phi(x)$, noise magnitude $D$, and the initial state distribution $p_0(x)$ so that the speed and accuracy of decisions in the model are similar to typical experimental values, and $f(x)$ is chosen to produce realistic firing rates\cite{Chandrasekaran:2017hp} (parameters provided in Supplementary Table~\ref{table1}). First, we infer the potential shape $\Phi(x)$ with the correct $D$ and $p_0(x)$ provided. For both ramping and stepping dynamics, our framework accurately infers the correct potential shape from 200 data trials (a realistic data amount in experiment, Fig.~\ref{fig4}a,b). At the iteration when the likelihoods of the fitted and ground-truth model are equal, the inferred potentials are in good agreement with the ground truth, confirming the inference accuracy. The inference accuracy further improves with a larger data amount of 1,600 trials.

Finally, we demonstrate simultaneous inference of all functions governing the non-stationary dynamics $\Phi(x)$, $p_0(x)$, and $D$ using synthetic data generated from the ramping model (Fig.~\ref{fig4}c). We update each of $\Phi(x)$, $p_0(x)$, and $D$ in turn on successive GD iterations. As the likelihood of the fitted model approaches the likelihood of the ground-truth model, the potential shape, noise magnitude, and the initial state distribution all closely match the ground truth, confirming the accurate inference of a full model of latent non-stationary Langevin dynamics.

Our framework accurately infers non-stationary Langevin dynamics from stochastic observations and can accommodate different observation processes, e.g., non-additive Poisson observation noise. We demonstrate that accurate inference requires taking into account the non-equilibrium initial and final states that represent the start and outcome of computations performed by the system. Ignoring the non-equilibrium initial or final states results in incorrect inference, in which erroneous features in the dynamics arise due to non-stationary data distribution. We illustrate our inference framework using models of neural dynamics during decision making, an inherently non-stationary process of transforming sensory information into a categorical choice. Comparisons between simple parametric models proved ineffective to reveal the underlying neural dynamics\cite{Latimer:2015hw,Chandrasekaran:2018dv,Zylberberg:2019el,Zoltowski:2019ig}. In contrast, our Langevin framework provides a flexible non-parametric description of dynamics, which can be used to directly infer the model from data. Our framework generalizes to several latent dimensions and parallel data streams\cite{genkin_mikhail_2020_4010952} (e.g., multi-neuron recordings) and opens new avenues for analyzing dynamics of complex systems far from equilibrium.

\section*{Methods}
 
\subsection{\label{sec:lc} Maximum-likelihood inference of latent non-stationary Langevin dynamics.}
We provide a brief summary of the analytical calculation of the model likelihood and its variational derivatives (see Supplementary Information for details). The likelihood $\mathscr{L}\left[Y(t) | \theta \right]$ is a conditional probability of observing the data $Y(t)$ given a model $\theta=\{\Phi(x),p_0(x),D\}$. The likelihood is obtained by marginalizing the joint probability density $P(\mathcal{X}(t),Y(t) | \theta) $ over all possible latent trajectories $\mathcal{X}(t)$ that may underlie the data\cite{Genkin:2020hz,Haas:2013gc}:
\begin{equation}\label{PI}
\mathscr{L} \left[ Y(t) |\theta \right] = \int \mathscr{D} \mathcal{X}(t) \; P(\mathcal{X}(t), Y(t) | \theta ).
\end{equation}
Here $\mathcal{X}(t)$ is a continuous latent trajectory, and the path integral is performed over all possible trajectories. We only consider here a single trial $Y(t)=\{t_0,t_1,...,t_N,t_E\}$, since the total data likelihood is a product of likelihoods of all trials. Using the Markov property of the latent Langevin dynamics Eq.~(\ref{E:LatentDynModel}) and conditional independence of spike observations, the joint probability density can be factorized (Fig.~\ref{fig1}b):
\begin{equation}\label{JP}
\begin{aligned}
P(X(t),Y(t))=p(x_{t_0})\left(\prod_{i=1}^{N}p(y_{t_i}|x_{t_i})p(x_{t_i}|x_{t_{i-1}})\right) p(x_{t_E}|x_{t_N})p(A|x_{t_E}).
\end{aligned}
\end{equation}
Here $p(x_{t_0})$ is the probability density of the initial latent state. $p(x_{t_i}|x_{t_{i-1}})$ is the transition probability density from $x_{t_{i-1}}$ to $x_{t_i}$ during the time interval between the adjacent spike observations. $p(y_{t_i}|x_{t_i})$ is the probability density of observing a spike at time $t_i$ given the latent state $x_{t_i}$. Finally, the term $p(A|x_{t_E})$ represents the absorption operator, which ensures that only trajectories terminating at one of the domain boundaries at time $t_E$ contribute to the likelihood. The absorption term $p(A|x_{t_E})$ is only applied in the case of absorbing boundaries, and it is absent in the case of reflecting boundaries (Supplementary Information~1).

The discretized latent trajectory $X(t)=\{x_{t_0},x_{t_1},...,x_{t_N},x_{t_E}\}$ is obtained by marginalizing the continuous trajectory $\mathcal{X}(t)$ over all latent paths connecting $x_{t_{i-1}}$ and $x_{t_{i}}$ during each interspike interval. These marginalizations are implicit in the transition probability densities. The likelihood is then calculated by marginalization over the discretized latent trajectory:
\begin{equation}\label{LM}
\mathscr{L}=\int_{x_{t_0}}\int_{x_{t_1}}\dots\int_{x_{t_N}} \int_{x_{t_E}}dx_{t_0} \dots dx_{t_E}  P(X(t),Y(t)).
\end{equation}

For the Langevin dynamics Eq.~(\ref{E:LatentDynModel}), the time-dependent probability density $\widetilde{p}(x,t)$ evolves according to the Fokker-Planck equation\cite{Risken:1996wg}:
\begin{eqnarray}\label{FP0}
\frac{\partial \widetilde{p}(x,t)}{\partial t} = \left(-D\frac{\partial}{\partial x}F(x)+D \frac{\partial^2}{\partial x^2}\right)\widetilde{p}(x,t)\equiv -\boldsymbol{\mathcal{\hat{H}}}_0\widetilde{p}(x,t),
\end{eqnarray}
which accounts for the drift and diffusion in the latent space (Fig.~\ref{fig2}, third column). In addition, the transition probability density $p(x_{t_i}|x_{t_{i-1}})$ in Eq.~(\ref{JP}) should also account for the absence of spike observations during intervals between adjacent spikes in the data. Thus, $p(x_{t_i}|x_{t_{i-1}})$ satisfies a modified Fokker-Planck equation:
\begin{eqnarray}\label{FP}
\frac{\partial p(x,t)}{\partial t} = \left(-D\frac{\partial}{\partial x}F(x)+D \frac{\partial^2}{\partial x^2} - f(x)\right)p(x,t)\equiv -\boldsymbol{\mathcal{\hat{H}}}p(x,t),
\end{eqnarray}
where the term $-f(x)$ accounts for the probability decay due to spike emissions\cite{Genkin:2020hz,Haas:2013gc}. The solution of this equation $p(x,t_{i})=p(x,t_{i-1})\exp(-\hat{\boldsymbol{\mathcal{H}}}(t_i-t_{i-1}))$ propagates the latent probability density forward in time during each interspike interval. Depending on the experiment design, we solve Eq.~(\ref{FP}) with either absorbing or reflecting boundary conditions (Supplementary Information~1). The probability densities of spike observations in Eq.~(\ref{JP}) are given by $p(y_{t_i}|x_{t_i})=f(x_{t_i})$ by the definition of the instantaneous Poisson firing rate. 

The term $p(A|x_{t_E})$ in Eq.~(\ref{JP}) represents the absorption operator $\boldsymbol{A}$, which ensures that the likelihood only includes trajectories terminating at the boundaries. The instantaneous probability $p_A$ for a trajectory to be absorbed at the boundaries given the latent state $x_{t_E}$ is obtained by applying $\boldsymbol{A}$ to a delta-function initial condition $\delta(x_{t_e})$ and then integrating over the latent space:
\begin{equation}\label{AOdef}
p_A=\int_{-1}^1\delta(x_{t_e})\boldsymbol{A}dx.
\end{equation}
To derive the absorption operator, we consider the survival probability $P_{\Delta t}(S_{t_E}|x_{t_E})$ for a trajectory to survive (i.e. not to be absorbed at the boundary) within a time interval $\Delta t$ given the latent state $x_{t_E}$. The survival probability is obtained by propagating the initial condition $\delta(x_{t_E})$ with the operator $\exp(-\hat{\boldsymbol{\mathcal{H}}}_0\Delta t)$ and integrating the result over the latent space:
\begin{equation}\label{SP}
P_{\Delta t}(S_{t_E}|x_{t_E})=\int_{-1}^{1}\delta(x_{t_E})\exp(-\hat{\boldsymbol{\mathcal{H}}}_0\Delta t)dx.
\end{equation}
Here we use the operator $\hat{\boldsymbol{\mathcal{H}}} _0$ instead of the operator $\hat{\boldsymbol{\mathcal{H}}}$, because the survival probability accounts only for the probability loss due to absorption at the boundaries and not for the probability decay due to spike emissions. 

The probability for a trajectory to be absorbed during a time interval $\Delta t$ given the state $x_{t_E}$ is given by $P_{\Delta t}(A_{t_E}|x_{t_E})=1-P_{\Delta t}(S_{t_E}|x_{t_E})$. Thus, the instantaneous probability of absorption is obtained as
\begin{eqnarray}\label{AP}
p_A=\lim\limits_{\Delta t\rightarrow 0} \frac{P_{\Delta t}(A_{t_E}|x_{t_E})}{\Delta t} =\lim\limits_{\Delta t\rightarrow 0} \frac{1-P_{\Delta t}(S_{t_E}|x_{t_E})}{\Delta t}=\int_{-1}^1 dx\delta(x_{t_E})\hat{\boldsymbol{\mathcal{H}}}_0,
\end{eqnarray}
where we use Eq.~(\ref{SP}) to take the limit. Comparing this result with Eq.~(\ref{AOdef}), we find that $\boldsymbol{A}=\hat{\boldsymbol{\mathcal{H}}}_0$. Note that $-\hat{\boldsymbol{\mathcal{H}}}_0$ is the Fokker-Planck operator in Eq.~(\ref{FP0}) that describes the rate of change of the latent probability density at each location $x$. Integrating both sides of Eq.~(\ref{FP0}) over the latent space, we obtain
\begin{equation}
    \frac{d \widetilde{p} (t)  }{d t}   =  \frac{d}{d t} \int_{-1}^1 dx \widetilde{p}(x,t)  = - \int_{-1}^1 dx \boldsymbol{A} \widetilde{p}(x,t).
\end{equation}
This equation describes the decay of the total probability $\widetilde{p}(t)=\int_x dx \widetilde{p}(x,t)$ in the latent space due to probability flux through the absorbing boundaries. Thus, applying the absorption operator $\boldsymbol{A}$ and integrating over the latent space represents the instantaneous loss of the total probability at time $t$, which is the fraction of all survived trajectories that reach the absorbing boundaries at exactly time $t$.

To compute and optimize the likelihood numerically, we represent Eq.~(\ref{LM}) in a discrete basis\cite{genkin_mikhail_2020_4010952} (Supplementary Information~1). In the discrete basis, all continuous functions, such as $p_0(x)$, are represented by vectors, and the transition, emission, and absorption operators are represented by matrices. Thus, Eq.~(\ref{LM}) is evaluated as a chain of vector-matrix multiplications.

\subsection{\label{sec:gd}Gradient descent optimization}
We minimize the negative log-likelihood with the gradient descent (GD) algorithm. Instead of directly updating the functions $\Phi(x)$ and $p_0(x)$, we, respectively, update the driving force $F(x)=-\Phi'(x)$ and an auxiliary function $F_0(x)\equiv p_0'(x)/p_0(x)$. The potential $\Phi(x)$ and $p_0(x)$ are obtained from $F(x)$ and $F_0(x)$ via
\begin{equation}\label{F0}
\Phi(x)=-\int_{-1}^xF(s)ds+C, \quad
p_0(x)=\frac{\exp\left(\int_{-1}^{x}F_0(s)ds\right)}{\int_{-1}^1\exp\left(\int_{-1}^{s'}F_0(s)ds\right)ds'}.
\end{equation}
We fix the arbitrary additive constant $C$ in the potential to satisfy $\int_{x}\exp[-\Phi(x)]dx=1$. The change of variable from $p_0(x)$ to $F_0(x)$ allows us to perform an unconstrained optimization of $F_0(x)$, and Eq.~(\ref{F0}) ensures that $p_0(x)$ satisfies the normalization condition for a probability density $\int_{-1}^1 p_0(x)dx=1$, $p_0(x)\geqslant 0$. We ensure the positiveness of the noise magnitude $D$ by rectifying its value after each GD update $D=\max(D,0)$. 

We derive analytical expressions for the variational derivatives of the likelihood $\delta\mathscr{L}/\delta{F(x)}$, $\delta\mathscr{L}/\delta{F_0(x)}$ and the derivative $\partial\mathscr{L}/\partial D$, which are then evaluated in the discrete basis (Supplementary Information~2,4). On each GD iteration, we update the model by stepping in the direction of the log-likelihood gradient:
\begin{equation}\label{GD}
	\Theta_{n+1}=\Theta_{n}+\gamma_\Theta\frac{1}{\mathscr{L}}\frac{\delta \mathscr{L}}{\delta \Theta}.
\end{equation}
Here $\Theta$ is one of the functions $F(x)$, $F_0(x)$, or the noise magnitude parameter $D$, with the corresponding learning rates $\gamma_\Theta>0$, and $n$ is the iteration number. For simultaneous inference of $\Phi(x)$, $p_0(x)$, and $D$ (Fig.~\ref{fig4}c), we update each of $F(x)$, $F_0(x)$, and $D$ in turn on successive GD iterations. The list of optimization hyperparameters, including learning rates and initializations, is provided in Supplementary Table~\ref{table1}.

\subsection{\label{sec:dg}Synthetic data generation}
To generate synthetic spike data from a model with given $\Phi(x)$, $p_0(x)$, and $D$, we numerically integrate Eq.~(\ref{E:LatentDynModel}) with the Euler–Maruyama method to produce latent trajectories $x(t)$ on each trial. We then use time-rescaling method\cite{Brown:2002jy} to generate spike times from an inhomogeneous Poisson process with the firing rate $\lambda(t)=f(x(t))$. We use 200 data trials in Fig.~\ref{fig3}; 200 and 1,600 trials in Fig.~\ref{fig4}a,b; and 400 trials in Fig.~\ref{fig4}c. These data amounts are typical for experiments in which neural activity is recorded during decision making\cite{Gold:2007fo,Chandrasekaran:2017hp}. A single experimental session usually contains a total of $1,000 - 2,000$ trails under different behavioral conditions, with about $100 - 200$ trials in each condition.

%
%

\section*{Acknowledgements}
This work was supported by the NIH grant R01 EB026949 (T.A.E. and M.G.), the Swartz Foundation (M.G.), and Katya H. Davey Fellowship (O.H.). We thank C. Aghamohammadi for thoughtful comments on the manuscript.

\section*{Author contributions}
M.G. and T.A.E. designed the research and developed the framework. M.G. and O.H. developed the code and performed computer simulations. M.G. and T.A.E. wrote the paper with input from O.H.

\section*{Competing interests}
The authors declare no competing interests.

\section*{Code availability}
The source code to reproduce results of this study will be made publicly available on GitHub
upon publication.

\section*{Additional Information}

Methods\\
Supplementary Information~1--4 \\
Supplementary Table~\ref{table1} \\
Supplementary References (\textit{32-34})


\newgeometry{textwidth=18.3cm,textheight=22cm}

\newpage

\newpage

\newpage

\newpage

\newgeometry{textwidth=16cm,textheight=21cm}

\newpage
\clearpage
\begin{center}
{\LARGE Supplementary Information for:\\ \vspace{0.3cm}
	Learning non-stationary Langevin dynamics from \\ \vspace{0.1cm} stochastic observations of latent trajectories}
	
\vspace{-0.1cm}
\large{Mikhail Genkin\textsuperscript{1}, Owen Hughes\textsuperscript{2}, and Tatiana A. Engel\textsuperscript{1}}
	
\vspace{-0.2cm}
\normalsize{\textsuperscript{1}Cold Spring Harbor Laboratory, Cold Spring Harbor, NY 11724}

\vspace{-0.5cm}
\normalsize{\textsuperscript{2}University of Michigan, Ann Arbor, MI}

\vspace{-0.5cm}
\normalsize{Corresponding author e-mail: engel@cshl.edu}
	
\end{center}
\setcounter{equation}{0}
\setcounter{figure}{0}
\setcounter{table}{0}
\setcounter{page}{1}
\makeatletter

\renewcommand{\theequation}{S\arabic{equation}}

\renewcommand{\cftsecleader}{\cftdotfill{\cftdotsep}}
\tableofcontents

\section{The likelihood calculation}\label{sec::appa}
The model likelihood is given by Eq.~(\ref{LM}), in which the transition probability densities $p(x_{t_i}|x_{t_{i-1}})$ are obtained from the solution of Eq.~(\ref{FP}), and the probability densities of spike observations are $p(y_{t_i}|x_{t_i})=f(x_{t_i})$. The absorption term $p(A|x_{t_E})$ only applies in the case of absorbing boundaries and is given by Eq.~(\ref{AP}). 

For an efficient numerical time-propagation of the latent probability density, we transform Eq.~(\ref{FP}) to the Hermitian form with the operator $\boldsymbol{\mathcal{H}}=\exp(\Phi(x)/2)\boldsymbol{\mathcal{\hat{H}}}\exp(-\Phi(x)/2)$, where $\boldsymbol{\mathcal{\hat{H}}}$ is defined in Eq.~(\ref{FP})\cite{Gardiner:1985wp}. The Hermitian operator $\boldsymbol{\mathcal{H}}$ evolves the scaled probability density $\rho(x,t)=p(x,t)\exp(\Phi(x)/2)$ according to:
\begin{equation}\label{appendix:HOE}
\frac{\partial \rho(x,t)}{\partial t}=-\boldsymbol{\mathcal{H}}\rho(x,t).
\end{equation}
This operator consists of two parts $\boldsymbol{\mathcal{H}}=\boldsymbol{\mathcal{H}}_0+\boldsymbol{\mathcal{H}}_I$, where $\boldsymbol{\mathcal{H}}_0$ accounts for the drift and diffusion in the latent space, and $\boldsymbol{\mathcal{H}}_I$ accounts for the decay of probability density due to spike emissions:
\begin{eqnarray}\label{appdix:H0HI}
\boldsymbol{\mathcal{H}}_0 &=& -e^{\Phi(x)/2}\frac{\partial}{\partial x} De^{-\Phi(x)}\frac{\partial}{\partial x} e^{\Phi(x)/2},
\nonumber \\
\boldsymbol{\mathcal{H}}_I &=& f(x).
\end{eqnarray}

We solve the eigenvector-eigenvalue problem for the Hermitian operator $\boldsymbol{\mathcal{H}}_0$:
\begin{eqnarray}\label{appdix:EVP1}
\boldsymbol{\mathcal{H}}_0\Psi_0(x)=\lambda\Psi_0(x).
\end{eqnarray}
Using Eq.~(\ref{appdix:H0HI}), we rewrite Eq.~(\ref{appdix:EVP1}) as:
\begin{eqnarray}\label{appdix:EVP}
-\frac{\partial}{\partial x} De^{-\Phi(x)}\frac{\partial}{\partial x} \phi_0(x) &=& \lambda_0 e^{-\Phi(x)}\phi_0(x).
\end{eqnarray}
Here $\phi_0(x)=\exp(\Phi(x)/2)\Psi_0(x)$ are the scaled eigenfunctions corresponding to the scaled probability density $\rho(x,t)\exp(\Phi(x)/2)=p(x,t)\exp(\Phi(x))$. The eigenfunctions of the operator $\boldsymbol{\mathcal{H}}_0$ are obtained as $\Psi_0(x)=\phi_0(x)\exp(-\Phi(x)/2)$.

For the case of absorbing boundaries, $p(x,t)$ satisfies the forward Fokker-Planck equation with absorbing boundary conditions
\begin{equation}\label{appdix:ABC}
\left[p(x,t)\right]_{x=\pm 1}=0,
\end{equation}
and the scaled probability density $p(x,t)\exp(\Phi(x))$ satisfies the backward Fokker-Planck equation also with absorbing boundary conditions\cite{Gardiner:1985wp}. Thus the eigenfunctions $\phi_0(x)$ are obtained by solving Eq.~(\ref{appdix:EVP}) with the boundary conditions $\phi_0(x)|_{x=\pm 1}=0$.

For the case of reflecting boundaries, $p(x,t)$ satisfies the Fokker-Planck equation with reflecting (zero flux) boundary conditions
\begin{equation}\label{appdix:RBC}
\left[Df(x)p(x,t)-D\frac{\partial p(x,t)}{\partial x} \right]_{x=\pm 1} = 0,
\end{equation}
and the scaled probability density satisfies the backward Fokker-Planck equation with zero derivative (Neumann) boundary conditions\cite{Gardiner:1985wp}. Thus the eigenfunctions $\phi_0(x)$ are obtained by solving Eq.~(\ref{appdix:EVP}) with the boundary conditions $\partial \phi_0 /\partial x |_{x=\pm 1} =0$.

To find the eigenfunctions of the operator $\boldsymbol{\mathcal{H}}$, we solve another eigenvalue-eigenvector problem $(\boldsymbol{\mathcal{H}}_0+\boldsymbol{\mathcal{H}}_I) \Psi(x) = \lambda \Psi(x)$ in the basis of the operator $\boldsymbol{\mathcal{H}}_0$.

As a result, we obtain an eigenbasis $\{\Psi_j(x)\}$ and eigenvalues $\{\lambda_j\}$ of the operator $\boldsymbol{\mathcal{H}}$, where $j=1,2,...N_v$ and the eigenbasis is truncated to the $N_v$ eigenvectors with the smallest eigenvalues. We refer to the operator $\boldsymbol{\mathcal{H}}$ as $\boldsymbol{\mathcal{H}}_{\rm abs}$ or $\boldsymbol{\mathcal{H}}_{\rm ref}$ for absorbing and reflecting boundary conditions, respectively. 

We solve the eigenvector-eigenvalue problem Eq.~(\ref{appdix:EVP}) using Spectral Elements Method (SEM)\cite{Genkin:2020hz,Haas:2013gc,Deville:2002uf}. We use the Gauss-Legendre-Lobatto (GLL) grid with Lagrange interpolation polynomials as the basis functions (SEM basis). In the SEM basis, each function is represented by a vector that consists of the function values evaluated at the SEM grid points $\{x_i\}$, where $i=1,2,...,N$. For example, the firing rate function $f(x)$ is represented by a vector $\boldsymbol{f}$ with components $f_i=f(x_i)$. In the SEM basis, the eigenbasis $\{\Psi_j(x)\}$ is a matrix $\boldsymbol{Q}$ of size $N \times N_v$ defined as $Q_{ij}=\Psi_j(x_i)$. The transformation between the SEM basis and the eigenbasis of $\boldsymbol{\mathcal{H}}$ is performed by generalized rules of basis transformations via $\boldsymbol{\tilde{f}}=\boldsymbol{Q^TW}\boldsymbol{f}$ and $\boldsymbol{f}=\boldsymbol{Q}\boldsymbol{\tilde{f}}$, where $\boldsymbol{\tilde{f}}$ is the representation of $\boldsymbol{f}$ in the eigenbasis of $\boldsymbol{\mathcal{H}}$. Here $\boldsymbol{W}={\rm diag}(\boldsymbol{w})$ is the SEM weight matrix that normalizes the matrix of eigenvectors: $ \boldsymbol{Q^TWQ}=\boldsymbol{I}$. Integration of functions in the SEM basis is performed with Gaussian quadrature by taking an inner product with the weight vector $\boldsymbol{w}$: $\int f(x) dx \approx \boldsymbol{f}^T\boldsymbol{w}$, where the weights for the GLL grid are calculated analytically\cite{Deville:2002uf}.

We compute the likelihood in the eigenbasis of $\boldsymbol{\mathcal{H}}$. The formal solution of Eq.~(\ref{appendix:HOE}) is given by
\begin{equation}
\rho(x_{t_k}|x_{t_{k-1}}) = e^{-\boldsymbol{\mathcal{H}} (t_k-t_{k-1})}\equiv e^{-\boldsymbol{\mathcal{H}} \Delta t_k}.
\end{equation}
In the eigenbasis of $\boldsymbol{\mathcal{H}}$, the corresponding matrix is diagonal:
\begin{equation}\label{SI:PrM}
T_{k,ij}=\int_x \Psi_i(x)e^{-\boldsymbol{\mathcal{H}\Delta t_k}} \Psi_j(x)dx = \delta_{ij}e^{-\lambda_i\Delta t_k}.
\end{equation}
In the $\boldsymbol{Q}$-basis, the spike emission operator matrix takes the form $\boldsymbol{E} = \boldsymbol{Q^T}\boldsymbol{W}\boldsymbol{f}\boldsymbol{Q}$. The absorption operator takes the form
$\boldsymbol{A}= \boldsymbol{Q}_{0H}^T{\rm diag}(\boldsymbol{\lambda}_0)\boldsymbol{Q}_{0H}$, where $\lambda_0$ are the eigenvalues of $\boldsymbol{\mathcal{H}}_0$ obtained from Eq.~(\ref{appdix:EVP1}), and $ \boldsymbol{Q}_{0H}$ is a transformation matrix from $\boldsymbol{\mathcal{H}}_0$-basis to $\boldsymbol{\mathcal{H}}$-basis. The integration over the terminal state $x_{t_E}$ in Eq.~(\ref{LM}) is realized by performing the inner product with a column vector $\boldsymbol{\beta}_{N+2}=\boldsymbol{Q}^T\boldsymbol{W\rho_{\rm eq}}$, where $\boldsymbol{\rho_{\rm eq}}$ is a representation of the function $\exp(-\Phi(x)/2)$ in the $\boldsymbol{\mathcal{H}}$ basis. Since the operator $\boldsymbol{\mathcal{H}}$ propagates the scaled probability density $\rho(x,t)$, first scaling $\rho(x,t)$ with $\exp(-\Phi(x)/2)$ and then taking the inner product with the weight vector corresponds to the integration of the original probability density $p(x,t)$.

Thus, in the $\boldsymbol{\mathcal{H}}$ basis, the likelihood Eq.~(\ref{LM}) is represented by a chain of matrix-vector multiplications:
\begin{equation}\label{appdix:L}
\mathscr{L} = \boldsymbol{\rho}_0^T \boldsymbol{T}_1 \boldsymbol{E} \boldsymbol{T}_2  \boldsymbol{E} ... \boldsymbol{T}_N  \boldsymbol{E}\boldsymbol{T}_{N+1}\boldsymbol{A} \boldsymbol{\beta}_{N+2},
\end{equation}
where the absorption operator $\boldsymbol{A}$ is applied only in the case of absorbing boundaries. Here $\boldsymbol{\rho}_0^T$ is a row vector that corresponds to the scaled initial probability density of latent states $p_0(x)\exp(\Phi(x)/2)$ in the eigenbasis of the operator $\boldsymbol{\mathcal{H}}$. The chain of matrix-vector multiplications in Eq.~(\ref{appdix:L}) realizes the integration over $x_{t_0}, x_{t_1},...,x_{t_N}$ in Eq.~(\ref{LM}). The last inner product with the vector $\boldsymbol{\beta}_{N+2}$ realizes the integration over the final $x_{t_E}$ in Eq.~(\ref{LM}).

On each iteration of the gradient descent, we evaluate the likelihood numerically by calculating the eigenvalues and the eigenfunctions of the operator $\boldsymbol{\mathcal{H}}$, computing the matrices $\boldsymbol{T}_k$, $\boldsymbol{E}$, $\boldsymbol{A}$, and performing the chain of matrix-vector multiplications in Eq.~(\ref{appdix:L}) from left to right (forward pass). To calculate the likelihood derivatives, we also perform a backward pass, where the chain in Eq.~(\ref{appdix:L}) is calculated in the reverse order.

\section{Variational derivatives of the likelihood}\label{sec::2}
The variational derivatives of the model likelihood are obtained using calculus of variations as described previously\cite{Genkin:2020hz}. The final analytical expressions read:
\begin{eqnarray}\label{appdix:LD}
\begin{aligned}
 \frac{\delta \mathscr{L}}{\delta F(x)} & =\sum_{ij} G_{ij} \frac{D}{2}e^{-\Phi(x)}\frac{d\left(\phi_i(x)\phi_j(x)\right)}{dx} +\frac{1}{2}\int_{-1}^x\left(\beta_0(s)p_0(s)e^{\Phi(s)/2}-\alpha_{N+2}(s)e^{-\Phi(s)/2}\right)ds, \\
 \frac{\delta \mathscr{L}}{\delta F_0(x)} & =\int_{-1}^x p_0(s)\left(\mathscr{L}-e^{\Phi(s)/2}\beta_0(s)\right)ds, \\
 \frac{\partial\mathscr{L}}{\partial D} & = -\int_{-1}^{1} dx e^{-\Phi(x)}\sum_{ij}G_{ij}\frac{d\phi_i(x)}{dx}\frac{d\phi_j(x)}{dx}.
\end{aligned}
\end{eqnarray}
Here $F_0(x)=p'_0(x)/p(x)$ is an auxiliary function used to perform unconstrained optimization of $p_0(x)$ (Methods). $\phi_i(x)=\Psi_i(x)\exp(\Phi(x)/2)$ are the scaled eigenfunctions of $\boldsymbol{\mathcal{H}}$.

The functions $\alpha_i(x)$ and $\beta_i(x)$ are calculated with, respectively, the forward and backward passes through the chain Eq.~(\ref{appdix:L}) in the eigenbasis of the operator $\boldsymbol{\mathcal{H}}$:
\begin{eqnarray}\label{appdix:alpha}
&&\boldsymbol{\alpha}_0 = \boldsymbol{\rho}_0^T, \nonumber \\
&&\boldsymbol{\alpha}_n =\boldsymbol{\alpha}_{n-1}  \boldsymbol{T}_n\boldsymbol{E}, \qquad n=1,2 \dots N, \nonumber \\
&&\boldsymbol{\alpha}_{N+1} = \boldsymbol{\alpha}_N \boldsymbol{T}_{N+1}, \quad  \boldsymbol{\alpha}_{N+2} = \boldsymbol{\alpha}_{N+1} \boldsymbol{A}.
\end{eqnarray} 
\begin{eqnarray}\label{appdix:beta}
&&\boldsymbol{\beta}_{N+1} = \boldsymbol{A\beta}_{N+2},  \nonumber \\
&&\boldsymbol{\beta}_n =  \boldsymbol{E}\boldsymbol{T}_{N+1}\boldsymbol{\beta}_{n+1}, \qquad n=1,2 \dots N, \nonumber \\
&&\boldsymbol{\beta}_0 = \boldsymbol{T}_{1}\boldsymbol{\beta}_{1}.
\end{eqnarray}
The absorption matrix $\boldsymbol{A}$ in Eqs.~(\ref{appdix:alpha}),(\ref{appdix:beta}) is applied only in the case of absorbing boundary conditions.

The matrix $\boldsymbol{G}$ is defined as:
\begin{equation}\label{SIG}
G_{ij} = \sum_{\tau=0}^{N+1}\Gamma_{ij}^{\tau+1} \alpha_{\tau,i}\beta_{\tau+1,j},
\end{equation}
where $i,j$ index the components of the vectors $\boldsymbol{\alpha}_\tau$ and $\boldsymbol{\beta}_{\tau+1}$. The matrix $\boldsymbol{\Gamma}^\tau$ is defined for $\tau=N+2$ as $\boldsymbol{\Gamma}^\tau=-\boldsymbol{I}$, and for $\tau<N+2$ it is defined as:
\begin{eqnarray}\label{appdix:gamma}
\Gamma^\tau_{ij}=\int_{0}^{\Delta t_\tau} e^{-(\Delta t_\tau-u)\lambda_i} e^{-u\lambda_j} du 
=\begin{cases}
\Delta t_\tau e^{-\lambda_i \Delta t_\tau}, \quad i=j, \\
\frac{e^{-\lambda_i \Delta t_\tau}-e^{-\lambda_j \Delta t_\tau}}{\lambda_j-\lambda_i}, \quad i\neq j,
\end{cases}
\end{eqnarray}
where $\Delta t_\tau=t_\tau-t_{\tau-1}$ are interspike intervals, and $\lambda_i$ are the eigenvalues of $\boldsymbol{\mathcal{H}}$.  The vectors $\alpha_i$ are calculated with a forward pass Eq.~(\ref{appdix:alpha}), and the vectors $\beta_i$ and $\boldsymbol{G}$ are subsequently calculated with a backward pass Eqs.~(\ref{appdix:beta})-(\ref{appdix:gamma}). The implementation of the forward and backward passes is based on the sum-product and scaling algorithms, similar to the Hidden Markov Models\cite{Bishop:2006ui}. The algorithm scales all $\boldsymbol{\alpha}_i$ and $\boldsymbol{\beta}_j$ so that $\alpha_i^T\beta_i = 1$ for all $i$, which prevents numerical underflow. The antiderivatives (indefinite integrals) in Eq.~(\ref{appdix:LD}) are evaluated by multiplication with the integration matrix that is constructed by analytical integration of the Lagrange interpolation polynomials\cite{Genkin:2020hz}.

While we perform computations in the SEM basis, the likelihood derivatives Eq.~(\ref{appdix:LD}) are expressed in terms of continuous functions, which can be discretized in any appropriate basis with arbitrary precision specified by the basis dimensionality. Detailed derivation of Eq.~(\ref{appdix:LD}) is provided in Supplementary Information~\ref{sec::4}.

\newpage

\section{Simulation parameters}   \label{sec::appd}
\begin{table*}[h]
	\begin{center}
		\begin{tabular}{| c | c | c | }
			\hline
			\hspace{1mm} \cellcolor[gray]{0.6}Parameter\hspace{1mm} 
			& \hspace{1mm} \cellcolor[gray]{0.6}Value \hspace{1mm} 
			& \hspace{1mm} \cellcolor[gray]{0.6}Description  \hspace{1mm} \\
			\hline
			\multicolumn{3}{|c|}{\cellcolor[gray]{0.8} Ground-truth model parameters}\\ \hline
			\hspace{1mm}$\Phi(x)$ \hspace{1mm} & \hspace{1mm} $-2.65x$
			\hspace{1mm} & \hspace{1mm} potential for ramping dynamics   \hspace{1mm} \\
			\hline
			\hspace{1mm} $\Phi(x)$ \hspace{1mm} & \hspace{1mm}  cubic splines*
			\hspace{1mm} & \hspace{1mm} potential for stepping dynamics \hspace{1mm}  \\
			\hline
			\hspace{1mm} $D$ \hspace{1mm} & \hspace{1mm} $0.56$
			\hspace{1mm} & \hspace{1mm} noise for ramping dynamics    \hspace{1mm} \\ 
			\hline
			\hspace{1mm}$D$ \hspace{1mm} & \hspace{1mm} $1$
			\hspace{1mm} & \hspace{1mm} noise for stepping dynamics \hspace{1mm} \\ \hline
			\hspace{1mm}$p_0(x)$ \hspace{1mm} & \hspace{1mm} $\propto\exp(-100x^2)$
			\hspace{1mm} & \hspace{1mm} initial probability density  \hspace{1mm} \\ \hline
			\hspace{1mm}$f(x)$ \hspace{1mm} & \hspace{1mm} $50x+60$
			\hspace{1mm} & \hspace{1mm} firing rate function, Hz \hspace{1mm} \\ 
			\hline

			\multicolumn{3}{|c|}{\cellcolor[gray]{0.8}Spectral Elements Method}\\ \hline
			\hspace{1mm}$x_{\rm begin}$ \hspace{1mm} & \hspace{1mm} -1
			\hspace{1mm} & \hspace{1mm} left boundary in the latent space \hspace{1mm} \\ \hline
			\hspace{1mm}$x_{\rm end}$ \hspace{1mm} & \hspace{1mm} 1
			\hspace{1mm} & \hspace{1mm}  right boundary in the latent space\hspace{1mm} \\ \hline
			\hspace{1mm} $N_v$ \hspace{1mm} & \hspace{1mm} 447
			\hspace{1mm} & \hspace{1mm}  number of eigenfunctions of $\boldsymbol{\mathcal{H}}$ \hspace{1mm} \\ \hline
			\hspace{1mm}$N_e$ \hspace{1mm} & \hspace{1mm} 64
			\hspace{1mm} & \hspace{1mm} number of elements \hspace{1mm} \\ \hline
			\hspace{1mm}$N_p$ \hspace{1mm} & \hspace{1mm} 8
			\hspace{1mm} & \hspace{1mm} number of grid points per element \hspace{1mm} \\ \hline
			
%
			\multicolumn{3}{|c|}{\cellcolor[gray]{0.8}Optimization hyperparameters}\\ \hline
			\hspace{1mm}$\gamma_F$ \hspace{1mm} & \hspace{1mm} $0.001\mbox{--}0.005$
			\hspace{1mm} & \hspace{1mm} learning rate for $F(x)$  \hspace{1mm} \\ \hline
			\hspace{1mm}$\gamma_D$ \hspace{1mm} & \hspace{1mm} 0.00025
			\hspace{1mm} & \hspace{1mm} learning rate for $D$ \hspace{1mm} \\ \hline
			\hspace{1mm}$\gamma_{F_0}$ \hspace{1mm} & \hspace{1mm} 0.025
			\hspace{1mm} & \hspace{1mm} learning rate for $F_0(x)$ \hspace{1mm} \\ \hline
			\hspace{1mm} $F^{(0)}(x)$ \hspace{1mm} & \hspace{1mm} 0
			\hspace{1mm} & \hspace{1mm} initialization for $F(x)$ \hspace{1mm} \\ \hline
			\hspace{1mm}$D^{(0)}$ \hspace{1mm} & \hspace{1mm} 1
			\hspace{1mm} & \hspace{1mm} initialization for $D$ \hspace{1mm} \\ \hline
			\hspace{1mm}$F^{(0)}_{0}(x)$ \hspace{1mm} & \hspace{1mm} 0
			\hspace{1mm} & \hspace{1mm} initialization for $F_0(x) $\hspace{1mm} \\ \hline
		\end{tabular}
	\end{center}
\caption{\textbf{Simulation parameters.}
*We define the potential function for stepping dynamics using six smoothly patched cubic splines. This potential shape is well approximated by the 14-degree polynomial: $\Phi(x) = 213.7 x^{14}  - 34.39 x^{13}  - 830.8 x^{12}  + 61.33 x^{11}  + 1329 x^{10}  + 37.88 x^9 - 1144 x^8 - 160.5 x^7 + 590.7 x^6 + 133 x^5 - 192.4 x^4 - 37.51 x^3 + 33.03 x^2 - 0.3233 x + 0.4446$.
}
	\label{table1}
\end{table*}

\section{Analytical derivation of the likelihood variational derivatives}\label{sec::4}
Here we derive the analytical expressions Eq.~(\ref{appdix:LD}). For convenience, we represent the likelihood using the Dirac bra-ket notation\cite{Genkin:2020hz,Haas:2013gc}:
\begin{equation}
\label{SI:LD}
\mathscr{L}\left[Y(t)|\theta\right]=\left \langle \rho_0 \middle|e^{-\boldsymbol{\mathcal{H}}( t_1-t_0)}\boldsymbol{y}_{t_1}e^{-\boldsymbol{\mathcal{H}}(t_2-t_1)}\boldsymbol{y}_{t_2}... e^{-\boldsymbol{\mathcal{H}}(t_N-t_{N-1})}\boldsymbol{y}_{t_N} e^{-\boldsymbol{\hat{\mathcal{H}}}(t_E-t_N)}\boldsymbol{A}\middle|\beta_{N+2} \right \rangle.
\end{equation}
In this notation, the scaled latent probability density $\rho(x,t)=p(x,t)\exp(\Phi(x)/2)$ is represented by the bra vectors, with the initial state $\langle \rho_0 |$. The time-propagation of the latent probability density is carried out by the operator $\exp(-\boldsymbol{\mathcal{H}}\Delta t_{\tau})$, where $\Delta t_\tau = t_\tau-t_{\tau-1}$. The operators $\boldsymbol{y}_{t_{\tau}}$ account for the observed spikes. The ket $| \beta_{N+2} \rangle$ accounts for the marginalization over the terminal latent state $x_{t_E}$.

To find variational derivative of the likelihood with respect to the force $F(x)$, we differentiate Eq.~(\ref{SI:LD}) applying the product rule of derivatives. The likelihood dependence on $F(x)$ is hidden inside the terms $\rho_0$ and $\beta_{N+2} $, and in each of the operators $\exp(-\boldsymbol{\mathcal{H}}\Delta t_{\tau})$, and $\boldsymbol{A}$. Thus, the derivative is given by the sum:
\begin{equation}\label{SI:deriv}
\begin{aligned}
\frac{\delta \mathscr{L}}{\delta F(x)}&=  \sum_{i,j} \left[\sum_{\tau=1}^{N+1} a_i(\tau-1) b_j(\tau) \frac{\delta \langle \Psi_i | e^{-\boldsymbol{\mathcal{H}}\Delta t_\tau}|\Psi_j \rangle}{\delta F} + a_i(N+1) b_j(N+2) \frac{\delta \langle \Psi_i | \boldsymbol{A}|\Psi_j \rangle}{\delta F} \right]+ \\ 
&+\sum_i\left[\frac{\delta a_i(0)}{\delta F} b_i(0)  + a_i(N+2) \frac{\delta b_i(N+2)}{\delta F}\right]\equiv L_1 +L_2,
\end{aligned}
\end{equation}
where $i$ and $j$ index the eigenfunctions of the operator $\boldsymbol{\mathcal{H}}$. In Eq.~(\ref{SI:deriv}), we introduced the quantities:
\begin{equation}\label{SIalpha}
\begin{aligned}
&\langle \alpha_0 | = \langle \rho_0 |, \qquad \langle\alpha_n |=\langle \alpha_{n-1} | e^{-\boldsymbol{\mathcal{H}}\Delta t_n}\boldsymbol{y}_{t_n},  \, n=1,2 \dots N+1, \\
&\langle \alpha_{N+1} | = \langle \alpha_{N} | e^{-\boldsymbol{\mathcal{H}}\Delta t_{N+1}}, \qquad \langle \alpha_{N+2} | = \langle \alpha_{N+1} | \boldsymbol{A}, \\
&| \beta_{N+1} \rangle = \boldsymbol{A} | \beta_{N+2} \rangle, \qquad | \beta_{N} \rangle = e^{-\boldsymbol{\mathcal{H}}\Delta t_{N+1}}| \beta_{N+1} \rangle, 
\\ &| \beta_{n} \rangle = \boldsymbol{y}_{t_n} e^{-\boldsymbol{\mathcal{H}}\Delta t_{n+1}}| \beta_{n+1} \rangle, \,  \, n=1,2, \dots N+1, \qquad 
| \beta_{0} \rangle = e^{-\boldsymbol{\mathcal{H}}\Delta t_{1}}| \beta_{1} \rangle, \\
&a_i(\tau)=\langle \alpha_\tau | \Psi_i \rangle, \qquad b_j(\tau)=\langle \Psi_j  | \beta_\tau \rangle .
\end{aligned}
\end{equation}
In a finite basis, such as the truncated eigenbasis of the operator $\boldsymbol{\mathcal{H}}$, each ket $\langle \alpha_\tau |$ corresponds to a vector $\boldsymbol{\alpha}_\tau$, and each $a_i(\tau)$ is the $i$-th entry of this vector, compare Eq.~(\ref{SIalpha}) with Eqs.~(\ref{appdix:alpha}),(\ref{appdix:beta}). Thus, $a_i(\tau)$ and $b_i(\tau)$ are calculated with the forward-backward pass.

We first derive the expression for the first term in Eq.~(\ref{SI:deriv}) denoted as $L_1$.
Using the formula for the derivative of an exponential operator\cite{Wilcox:1967jq}, we obtain\cite{Genkin:2020hz,Haas:2013gc}:
\begin{equation}\label{SI:expder}
\begin{aligned}
\frac{\delta \langle \Psi_i | e^{-\boldsymbol{\mathcal{H}}\Delta t_\tau}|\Psi_j \rangle}{\delta F} =  - \left \langle \Psi_i \middle | \frac{\partial \boldsymbol{\mathcal{H}}}{\partial F} \middle|\Psi_j \right \rangle \Gamma^\tau_{i,j},
\end{aligned}
\end{equation}
where the matrix $\boldsymbol{\Gamma}^\tau$ is defined in Eq.~(\ref{appdix:gamma}). The operator $\boldsymbol{\mathcal{H}}$ can be written as\cite{Haas:2013gc}:
\begin{equation}\label{appdix:H}
\boldsymbol{\mathcal{H}}=-D\nabla^2+D\frac{F'(x)}{2}+D\frac{F^2(x)}{4}-f(x).
\end{equation}
Using the Euler-Lagrange equation for a variational derivative, we obtain:
\begin{equation}\label{appdix:Hd}
\begin{aligned}
\left \langle \Psi_i \middle | \frac{\delta \boldsymbol{\mathcal{H}}}{\delta F} \middle|\Psi_j \right \rangle =
\frac{\delta}{\delta F}D\int \Psi_i(x) \left(\frac{F'(x)}{2}+\frac{F(x)^2}{4}\right)\Psi_j(x)dx = \\
\frac{D}{2}\left(F(x)\Psi_i(x)\Psi_j(x)-\frac{d\left(\Psi_i(x)\Psi_j(x)\right)}{dx}\right)=-\frac{D}{2}\exp(-\Phi(x))\frac{d(\phi_i(x)\phi_j(x))}{dx}.
\end{aligned}
\end{equation}
Here we normalize $\Phi(x)$ such that $\int \exp(-\Phi(x))dx=1$, and $\phi_i(x)=\Psi_i(x)\exp(\Phi(x)/2)$ are the scaled eigenfunctions.

The absorption operator is $\boldsymbol{A}=\boldsymbol{\mathcal{H}}_0$, where $\boldsymbol{\mathcal{H}}_0$ is the part of $\boldsymbol{\mathcal{H}}$ that accounts for the drift and diffusion in the latent space:
\begin{equation}\label{appdix:H0}
\boldsymbol{\mathcal{H}}_0=-D\nabla^2+D\frac{F'(x)}{2}+D\frac{F^2(x)}{4}.
\end{equation}
Therefore, the derivative of the operator $\boldsymbol{A}$ with respect to $F(x)$ is the same as the derivative of the operator $\boldsymbol{\mathcal{H}}$ and is given by Eq.~(\ref{appdix:Hd}). 

Combining Eqs.~(\ref{SI:expder}),(\ref{appdix:Hd}), we obtain the expression for $L_1$ (c.f. with the first term in Eq.~(\ref{appdix:LD})):
\begin{equation}
L_1=\sum_{ij} G_{ij} \frac{D}{2}e^{-\Phi(x)}\frac{d\left(\phi_i(x)\phi_j(x)\right)}{dx}.
\end{equation}
Here the matrix $\boldsymbol{G}$, defined in Eq.~(\ref{SIG}), gathers the contributions from all terms $\exp(-\boldsymbol{\mathcal{H}}\Delta t_i)$ and $\boldsymbol{A}$ (where for $\tau=N+2$ we define $\boldsymbol{\Gamma}^\tau=-\boldsymbol{I}$, so that Eq.~(\ref{SI:expder}) also holds for the operator $\boldsymbol{A}$).

Next, we compute the second term in Eq.~(\ref{SI:deriv}) denoted as $L_2$, which requires calculating the derivatives of $a_i(0)$ and $b_i(N+2)$ with respect to $F(x)$. For convenience, we write this term in the function representation:
\begin{equation}\label{appdix:L2}
\begin{aligned}
L_2&=\sum_i\left[\frac{\delta a_i(0)}{\delta F} b_i(0)  + a_i(N+2) \frac{\delta b_i(N+2)}{\delta F}\right] = \\ &=\frac{\delta}{\delta F(x)}\left[\int_{-1}^1 \alpha_0(x)\beta_0(x)dx+\int_{-1}^1 \alpha_{N+2}(x)\beta_{N+2}(x)dx\right]\equiv \frac{\delta}{\delta F(x)} \left[l_1+l_2\right] .
\end{aligned}
\end{equation}
Here the functions $\alpha_i(x)$ and $\beta_i(x)$ correspond to $\langle \alpha_i |$ and $| \beta_i \rangle$ in Eq.~(\ref{SIalpha}), respectively. We have $\alpha_0(x)=p_0(x)\exp(\Phi(x)/2)$, $\beta_{N+2}(x)=\exp(-\Phi(x)/2)$ (Methods), and the derivative does not apply to the functions $\beta_0(x)$ and $\alpha_{N+2}(x)$, since their dependence on $F(x)$ is already accounted for in the term $L_1$ in Eq.~(\ref{SI:deriv}). The derivatives are evaluated using the chain rule for functional derivatives:
\begin{equation}
\frac{\delta l_1}{\delta F(x)} = \int_{-1}^1  ds\frac{\delta l_1}{\delta \Phi(s)} \frac{\delta \Phi(s)}{\delta F(x)},
\end{equation}
where $l_1=\int_{-1}^1 p_0(x)\exp(\Phi(x)/2)\beta_0(x)dx$. Using the Euler-Lagrange equation, we obtain:
\begin{equation}
\begin{aligned}
\frac{\delta l_1}{\delta \Phi(s)}&=\frac{1}{2}p_0(s)\exp(\Phi(s)/2)\beta_0(s), \\
\frac{\delta \Phi(s)}{\delta F(x)}&= -\frac{\delta}{\delta F(x)}\int_{-1}^sF(s')ds'= - \frac{\delta}{\delta F(x)}\int_{-1}^1F(s')H(s-s')ds'= H(x-s),
\end{aligned}
\end{equation}
where $H$ is the Heaviside step function. Thus, we obtain:
\begin{equation}
\frac{\delta l_1}{\delta F(x)}=\int_{-1}^1ds \frac{1}{2}p_0(s)\exp(\Phi(s)/2)b_0(s)H(x-s)= \frac{1}{2} \int_{-1}^xds p_0(s)\exp(\Phi(s)/2)b_0(s) .
\end{equation}
The variational derivative of the second term $l_2$ in Eq.~(\ref{appdix:L2}) is calculated similarly. As a result, we obtain the expression for $L_2$ (c.f. with the second term in Eq.~(\ref{appdix:LD})):
\begin{equation}
L_2 = \frac{1}{2}  \int_{-1}^xds \left(p_0(s)\exp(\Phi(s)/2)b_0(s) - a_{N+2}(s) \exp(-\Phi(s)/2)\right).  
\end{equation}

To derive the expression for $\delta \mathscr{L}/\delta F_0(x)$, we write the likelihood in the function representation:
\begin{equation}
\mathscr{L}=\int_{-1}^{1} p_0(x)\exp(\Phi(x)/2)\beta_0(x)dx.
\end{equation}
The likelihood depends on $F_0(x)$ only through the term $p_0(x)$:
\begin{equation}\label{appdix:F0d}
\frac{\delta\mathscr{L}}{\delta F_0(x)}=\int_{-1}^1 ds \frac{\delta\mathscr{L}}{\delta p_0[s]}\frac{\delta p_0[s]}{\delta F_0(x)}.
\end{equation}
We compute the derivative $\delta\mathscr{L}/\delta p_0[s]$ using the Euler-Lagrange equation:
\begin{equation}\label{appdix:p0d2}
\frac{\delta\mathscr{L}}{\delta p_0[s]} = \exp(\Phi(s)/2)\beta_0(s).
\end{equation}
To compute the derivative $\delta p_0[s]/\delta F_0(x)$, we express $p_0(x)$ through $F_0(x)$ (Methods):
\begin{equation}
p_0(x)=\frac{\exp(\int_{-1}^x F_0(x')dx')}{\int_{-1}^1\exp(\int_{-1}^x F_0(x')dx')dx} \equiv \frac{e^{G(x)}}{\int_{-1}^1e^{G(x')}dx'}= \frac{\int_{-1}^1 e^{G(x')}\delta(x-x')dx'}{\int_{-1}^1e^{G(x')}dx'}.
\end{equation}
Here $G(x)=\int_{-1}^1F_0(x')H(x-x')dx'$, and $H$ is the Heaviside step function. Using the chain rule, we obtain:
\begin{equation}
\frac{\delta p_0[s]}{\delta F_0(x)}=\int_{-1}^1ds'\frac{\delta p_0[s]}{\delta G[s']}\frac{\delta G[s']}{\delta F_0(x)}.
\end{equation}
Using the Euler-Lagrange equation, we get:
\begin{equation}
\begin{aligned}
&\frac{\delta p_{0}[s]}{\delta G[s']}= \frac{e^{G(s')}\delta(s-s')\int_{-1}^1 e^{G(x')}dx'-e^{G(s')}e^{G(s)}}{\left(\int_{-1}^1 e^{G(x')}dx'\right)^2}, \\
&\frac{\delta G[s']}{\delta F(x)} = H(s'-x).
\end{aligned}
\end{equation}
As a result,
\begin{equation}\label{SIp2}
\begin{aligned}
\frac{\delta p_0[s]}{\delta F_0(x)} & =\int_{-1}^1ds' \left[\frac{e^{G(s')}\delta(s-s')}{\int_{-1}^1 e^{G(x')}dx'} -\frac{e^{G(s')}e^{G(s)}}{\left(\int_{-1}^1 e^{G(x')}dx'\right)^2}\right] H(s'-x) =
\\
&= \frac{e^{G(s)}}{\int_{-1}^1 e^{G(x')}dx'}\left[H(s-x)-\frac{\int_{-1}^1e^{G(s')}H(s'-x)ds'}{\int_{-1}^1 e^{G(x')}dx'}\right] =
\\
& = p_0(s) \left[H(s-x)-\frac{\int_{x}^1e^{G(s')ds'}}{\int_{-1}^1 e^{G(x')}dx'}\right] =
\\
&= p_0(s) \left[H(s-x)-\int_x^1p_{0}(s')ds'\right]  = p_{0}(s) \left[-H(x-s) + \int_{-1}^xp_{0}(s')ds'\right].
\end{aligned}
\end{equation}
By substituting Eqs.~(\ref{SIp2}),(\ref{appdix:p0d2}) into Eq.~(\ref{appdix:F0d}) we obtain:
\begin{equation}\label{SI:dF0}
\begin{aligned}
\frac{\delta\mathscr{L}}{\delta F_0(x)}=\int_{-1}^1 ds \exp(\Phi(s)/2)\beta_0(s)p_{0}(s) \left[-H(x-s) + \int_{-1}^xp_{0}(s')ds'\right] = \\
=-\int_{-1}^x ds \exp(\Phi(s)/2)\beta_0(s)p_{0}(s) + \int_{-1}^xp_{0}(s')ds' = \int_{-1}^{x} p_0(s) \left(\mathscr{L}-\exp(\Phi(s)/2)\beta_0(s)\right)ds.
\end{aligned}
\end{equation}

Finally, we derive the expression for the derivative $d\mathscr{L}/dD$. The likelihood Eq.~(\ref{SI:LD}) depends on $D$ through the operators $\exp(-\boldsymbol{\mathcal{H}}\Delta t_{\tau})$ and $\boldsymbol{A}$. Applying the product rule for derivatives, we obtain:
\begin{equation}\label{SI:Dder}
\frac{\partial \mathscr{L}}{\partial D} =  \sum_{i,j} \left[\sum_{\tau=1}^{N+1} a_i(\tau-1) b_j(\tau) \frac{ \partial \langle \Psi_i | e^{-\boldsymbol{\mathcal{H}}\Delta t_\tau}|\Psi_j \rangle}{\partial D} + a_i(N+1) b_j(N+2) \frac{\partial \langle \Psi_i | \boldsymbol{A}|\Psi_j \rangle}{\partial D} \right],   
\end{equation}
where $a_i(\tau)$, $b_j(\tau)$ are defined in Eq.~(\ref{SIalpha}). Similarly to Eq.~(\ref{SI:expder}), we have
\begin{equation}\label{SI:expder2}
\begin{aligned}
\frac{\partial \langle \Psi_i | e^{-\boldsymbol{\mathcal{H}}\Delta t_\tau}|\Psi_j \rangle}{\partial D} =  - \left \langle \Psi_i \middle | \frac{\partial \boldsymbol{\mathcal{H}}}{\partial D} \middle|\Psi_j \right \rangle \Gamma^\tau_{i,j},
\end{aligned}
\end{equation}
where the matrix $\Gamma^\tau$ is defined in Eq.~(\ref{appdix:gamma}). Using Eq.~(\ref{appdix:H}) and the Euler-Lagrange equation, we obtain:
\begin{equation}\label{appdix:Dd}
\begin{aligned}
\left \langle \Psi_i \middle | \frac{d \boldsymbol{\mathcal{H}}}{d D} \middle|\Psi_j \right \rangle = \int_{-1}^{1} \left[ \Psi_i(x)\left(\frac{F'(x)}{2}+\frac{F^2(x)}{4}\right)\Psi_j(x)-\Psi_i(x)\Psi_j^{''}(x) \right]dx = \\
-\int_{-1}^{1} \left[\phi_i(x)\left(\exp(-\Phi(x))\phi_j'(x)\right)'\right]dx = \int_{-1}^{1} \exp(-\Phi(x)) \phi_i'(x)\phi_j'(x)dx.
\end{aligned}
\end{equation}
Combining Eqs.~(\ref{SI:Dder}),(\ref{SI:expder2}),(\ref{appdix:Dd}), we obtain (c.f. Eq.~(\ref{SI:deriv})):
\begin{equation}
\frac{\partial\mathscr{L}}{\partial D}  = -\int_{-1}^{1} dx e^{-\Phi(x)}\sum_{ij}G_{ij}\phi_i'(x)\phi_j'(x),    
\end{equation}
where the matrix $\boldsymbol{G}$ is defined by Eq.~(\ref{SIG}).

\let\oldthebibliography=\thebibliography
\let\oldendthebibliography=\endthebibliography
\renewenvironment{thebibliography}[1]{%
	\oldthebibliography{#1}%
	\setcounter{enumiv}{31}%
}{\oldendthebibliography}

\end{document}